\documentclass{article}
\usepackage{spconf,amsmath,graphicx}
\usepackage[most]{tcolorbox}%
\usepackage{times}
\usepackage{latexsym}
\usepackage{amsmath} %for \boldsymbol
\usepackage{microtype}
\usepackage{booktabs}
\usepackage{multirow}
\usepackage{bbm}
\usepackage{blindtext}
\usepackage{hyperref}
\title{Improving Open Information Extraction with Large Language Models: A Study on Demonstration Uncertainty}
\begin{document}
%\name{Chen Ling$^1$, Xujiang Zhao$^2$, Xuchao Zhang$^3$, Yanchi Liu$^2$, Wei Cheng$^2$, Haoyu Wang$^2$, \\ Zhengzhang Chen$^2$, Takao Osaki$^4$, Katsushi Matsuda$^4$, Haifeng Chen$^2$, Liang Zhao$^1$}
\name{%
\begin{tabular}{@{}c@{}}
Chen Ling $^{1, 2}$ \quad 
Xujiang Zhao$^{2}$ \quad 
Xuchao Zhang$^{3}$ \quad
Yanchi Liu$^2$ \quad
Wei Cheng$^2$ \quad
Haoyu Wang$^2$ \\ 
Zhengzhang Chen$^2$ \quad
Takao Osaki$^4$ \quad
Katsushi Matsuda$^4$ \quad 
Haifeng Chen$^2$ \quad 
Liang Zhao$^{1}$
\end{tabular}}
\address{$^1$Emory University,
$^2$NEC Labs,
$^3$Microsoft,
$^4$NEC Corporation
}
\maketitle
\begin{abstract}
Open Information Extraction (OIE) task aims at extracting structured facts from unstructured text, typically in the form of (subject, relation, object) triples. Despite the potential of large language models (LLMs) like ChatGPT as a general task solver, they lag behind state-of-the-art (supervised) methods in OIE tasks due to two key issues. First, LLMs struggle to distinguish irrelevant context from relevant relations and generate structured output due to the restrictions on fine-tuning the model. Second, LLMs generates responses autoregressively based on probability, which makes the predicted relations lack confidence. In this paper, we assess the capabilities of LLMs in improving the OIE task. Particularly, we propose various in-context learning strategies to enhance LLM's instruction-following ability and a demonstration uncertainty quantification module to enhance the confidence of the generated relations. Our experiments on three OIE benchmark datasets show that our approach holds its own against established supervised methods, both quantitatively and qualitatively. 
The code and data can be found at: \url{https://github.com/lingchen0331/demonstration_uncertainty}.
\end{abstract}
\section{Introduction}
Open Information Extraction (OIE) \cite{zhou2022survey} involves the identification and extraction of novel relations and their components (e.g., subjective, action, objective, and adverbs) from unstructured text. It enables the creation of large-scale knowledge graphs from diverse sources \cite{wang2018information}, aiding in tasks like question answering \cite{ling2023knowledge}, knowledge-augmented reasoning \cite{chowdhury2023knowledge}, and semantic search \cite{niklaus2018survey}. As a frontier technology, ChatGPT \cite{ouyang2022training} and other large language models (LLMs) \cite{ling2023beyond} excel at comprehending and producing a wide variety of intricate natural language constructs. Therefore, they naturally present a promising solution for solving the OIE task without the need for substantial training. 

The dominant OIE methods \cite{yu2021maximal,zhan2020span,kolluru2020imojie} are trained on labeled data, where each entity and their relations are explicitly annotated. This allows them to learn precise patterns and directly map input to specific output tags, resulting in high accuracy. Despite the potential of LLMs like ChatGPT as a general task solver, they lag behind tagging-based methods in OIE tasks due to two key issues \cite{ling2023beyond}. First, LLMs as a generative model are trained to generate human-like text and not specifically for information extraction. While they have a broad understanding of language and can generate coherent responses, they may not be as accurate or consistent in extracting specific pieces of information from the text as supervised models trained specifically for that task. Second, the responses generated by LLMs are based on the input prompt and are probabilistic in nature, which can result in outputs with lower confidence. This lack of confidence can engender inconsistencies, such as the same relation being extracted differently in varying contexts or not being extracted at all in certain instances. Furthermore, this diminished confidence can lead to the extraction of incorrect or irrelevant relations, thereby reducing confidence in interpreting the extracted relations.

\begin{figure*}[t]
\centering
\includegraphics[width=\textwidth]{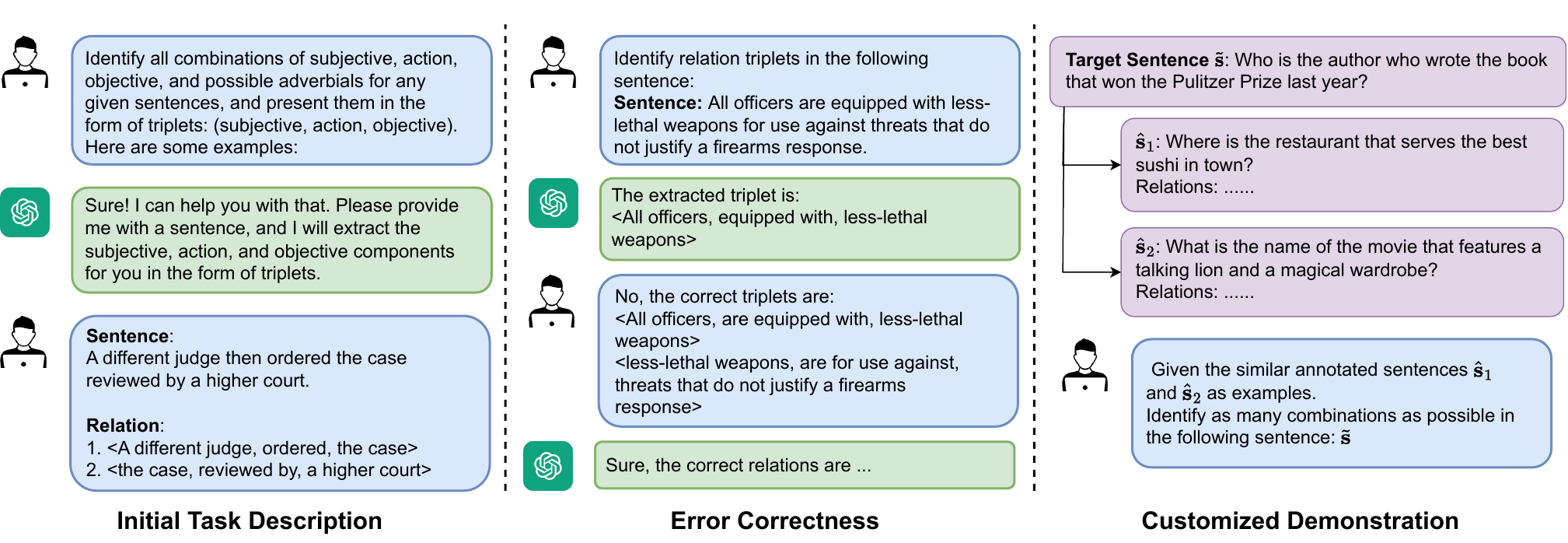}
\vspace{-3mm}
\caption{The framework of the proposed method consists of 1) providing an initial task description; 2) setting up a quiz to enhance ChatGPT's understanding of the OIE task, and 3) customized demonstration selection.}
\label{fig: example}
\end{figure*}

While zero-shot LLMs cannot solve complex OIE problems solely with original task instructions \cite{ling2023beyond}, there are a few attempts \cite{lu2023pivoine,han2023information,wei2023zero} trying to solve OIE with LLMs in different ways. A recent method \cite{wei2023zero} is proposed to tackle the information extraction with ChatGPT in an interactive manner by decomposing the framework into several parts and then combining the results of each round into a final structured result, but they can only handle OIE tasks with fixed relations. Lu et al. also proposed to leverage instruction tuning to enhance LLMs on a hand-crafted dataset, however, their method has to be extensively fine-tuned on hand-crafted datasets \cite{lu2023pivoine}. Another recent approach \cite{han2023information} focuses on investigating the capability of ChatGPT in the OIE task from various aspects. However, none of the existing works have considered enhancing the robustness and confidence of the response.

We summarize the contributions of our work as follows: (1) We propose a novel framework that allows LLMs to solve OIE tasks with various few-shot demonstration strategies without extensive fine-tuning; (2) We include an uncertainty quantification module to increase the confidence of the predicted answers. (3) A series of experiments have shown the effectiveness of the proposed method as well as each component in the proposed framework.

\section{Approach}
In this work, our primary focus is on the task of Relational Triplet Extraction, which is designed to identify entities and their relationships from a given sentence. To enable LLMs to produce responses that meet the requirements of the task, we employ a strategy that incorporates task-specific instructions, few-shot demonstrations, and an error correction mechanism. This approach facilitates the generation of coherent and structured responses. Moreover, to counteract the potential drawbacks associated with low confidence in the generated answers, we incorporate an demonstration uncertainty quantification mechanism. This mechanism serves to filter out relational triplets that exhibit a high degree of uncertainty, thereby enhancing the output reliability.

\subsection{Relation Extraction with In-context Learning}
\noindent\textbf{Problem Formulation.} Formally, given a sentence as a sequence of tokens/words $\Tilde{\mathbf{s}}=<w_1,w_2, \cdots, w_n>$, Relational Triplet Extraction requires to output a list of tuples $T = [T_1, T_2, \cdots]$ with the $i$-th tuple $T_i = 〈w_s, p_i, w_o>$ representing a fact in the source sequence, where $p$ denotes the predicate in $T_i$, $w_s$ and $w_o$ are the subjective and objective entities of $T_i$, respectively. As an autogressive generation model, ChatGPT outputs $T$ based on the input prompt (i.e., task instruction), and the response is directly determined by designing a suitable prompt/instruction. 

\noindent\textbf{Initial Instruction Crafting.} Task-specific instructions are essential for Open Information Extraction, particularly for the Relational Triplet Extraction task, as they guide the model to navigate the complexity and ambiguity of unstructured natural language and output structured responses for the task. Specifically, we incorporate a unified workflow with chain of instructions to guide the model step-by-step. As demonstrated in Figure \ref{fig: example}, ChatGPT is kick-started by prompted with an \textit{Initial Task Description} and a few demonstrations.  

\noindent\textbf{Error Correction.} The error correction mechanism refines ChatGPT's understanding of the OIE task and enhances response accuracy. After introducing easy examples to ChatGPT, a quiz, acting as a validation tool, is set. As depicted in Figure \ref{fig: example}, ChatGPT is tasked with extracting relational triplets from several sentences without prior correct annotations. Subsequent provision of correct answers helps rectify potential errors or incomplete responses. Empirically, this step bolsters ChatGPT's performance on actual OIE tasks, improving accuracy and dependability.

\noindent\textbf{Few-shot Demonstration Selection.} To better enable ChatGPT generates responses consistent with the task description, we incorporate few-shot examples sampled from the training set that are similar to the target sentence $\Tilde{\mathbf{s}}$. Specifically, given a training set $\mathcal{S}$ consisting of annotated sentences, we aim to retrieve a small subset $\hat{\mathcal{S}} \subseteq \mathcal{S}$ of structurally similar sentences with the target sentence $\Tilde{\mathbf{s}}$. For example, if the target sentence $\Tilde{\mathbf{s}}$ is an interrogative with an attributive clause: \texttt{Who is the author who wrote the book that won the Pulitzer Prize last year?}, then the ideal structurally-similar sentences from the training set should be: \texttt{Where is the restaurant that serves the best sushi in town?}. Each testing instance is given customized in-context learning demonstrations rather than fixed ones. Structural similarity is computed using cosine similarity between sentence latent embeddings, obtained through an Instruction-Fine-tuned Language Model \cite{su2022one}.

\begin{table*}[t]\centering
\resizebox{0.86\textwidth}{!}{%
\begin{tabular}{@{}lccccccccc@{}}
\toprule
\multirow{2}{*}{}                             & \multicolumn{3}{c}{CaRB} & \multicolumn{3}{c}{OIE2016} & \multicolumn{3}{c}{ReOIE}                     \\ \cmidrule(l){2-10} 
                                              & PR     & RE     & F1     & PR      & RE     & F1     & PR            & RE            & F1            \\ \midrule
\textbf{Rule-based}: OpenIE4                           & -      & -      & 48.0   & -       & -      & 60.0   & -             & -             & 68.3          \\
\textbf{Tagging-based}: SpanOIE                        & \underline{60.9}   & 41.6   & 49.4   & -       & -      & \textbf{69.4}   & \textbf{79.7} & \underline{74.5}          & \textbf{77.0} \\
\textbf{Generation-based}: IMOJIE                      & \textbf{64.7}   & 45.6   & \textbf{56.8}   & -       & -      & -      & \underline{88.1}          & 67.1          & \underline{76.2}          \\ \midrule
\textsc{LLaMA-2-13B w/ Fixed Demo}                                 & 41.6   & 28.0   & 33.4  & 37.5    & 30.6   & 33.7   & 31.2         & 18.1          & 22.9          \\
\textsc{LLaMA-2-13B w/ Selected Demo}                                 & 43.2   & 30.4   & 35.7  & 40.7    & 32.6   & 36.2   & 47.9         & 34.5          & 40.1          \\
\textsc{LLaMA-2-13B w/ Selected Demo \& Uncertainty}                                 & 44.4   & 30.6   & 36.2  & 41.1    & 33.5   & 36.9   & 32.3         & 21.3          & 25.7          \\\midrule
\textsc{LLaMA-2-70B w/ Fixed Demo}                                 & 43.7   & 51.0   & 47.1  & 57.5    & \textbf{71.3}   & 63.7   & 60.9          & 70.4          & 65.3          \\
\textsc{LLaMA-2-70B w/ Selected Demo}                                 & 55.4   & 47.1   & 50.9  & \underline{68.4}    & 61.6   & 64.8   & 62.1         & 72.5          & 66.9          \\
\textsc{LLaMA-2-70B w/ Selected Demo \& Uncertainty}                                 & 56.3   & 47.5   & 51.5  & \textbf{68.9}    & 63.0   & \underline{65.8}  & 69.4         & 65.7          & 67.5          \\
\midrule
\textsc{GPT-3.5-Turbo}                                 & 41.0   & 37.5   & 39.1  & 41.5    & 51.3   & 45.9   & 35.8          & 20.2          & 25.9          \\
\textsc{GPT-3.5-Turbo w/ Fixed Demo}                   & 53.8   & 48.6   & 51.1   & 60.6    & 64.5   & 62.5   & 64.5          & 71.4          & 67.8          \\
\textsc{GPT-3.5-Turbo w/ Selected Demo}                & 54.7   & \underline{49.5}   & 52.0   & 59.5    & 67.0   & 63.1   & 50.5          & \textbf{79.2} & 61.6          \\
\textsc{GPT-3.5-Turbo w/ Selected Demo \& Uncertainty} & 54.3   & \textbf{50.1}   & \underline{52.1}   & 61.6    & \underline{69.1}   & 65.1   &       65.8        &       70.2        &         67.9      \\ \bottomrule
\end{tabular}%
}\vspace{-3mm}
\caption{The performance of each approach on three popular benchmarks CaRB, OIE2016, and ReOIE with multiple partial matching strategies. The best results is highlighted with bold and the second best is highlighted with underline. The results missing in the literature are marked as -.}
\vspace{-3mm}
\label{tab:my-table}
\end{table*}

\subsection{Demonstration Uncertainty}\label{sec:demo}
After completing the in-context training steps with few-shot demonstration selections and error correctness, the performance can already go beyond the original zero-shot ChatGPT. However, the autoregressive generation process's inherent probabilistic characteristics may impair confidence in the predictions. In this study, we introduce an innovative module designed to quantitatively assess the uncertainty associated with the predicted output, thereby bolstering confidence in the  output.

In this work, we leverage an ensemble method to quantify the uncertainty in ChatGPT's responses. Specifically, after obtaining the structural similar annotated sentence set $\hat{\mathcal{S}} \subseteq \mathcal{S}$, we sample a list of small subset $[\hat{\mathcal{S}}_i], \hat{\mathcal{S}}_i \sim \hat{\mathcal{S}}$. We iteratively allow ChatGPT to generate answers of the target sentence $\Tilde{\mathbf{s}}$ by employing $\hat{\mathcal{S}}_i$ as an in-context learning example. Moreover, we leverage the prompt: ``\texttt{Identify \textbf{as many combinations as possible} in the following sentence:} $\Tilde{\mathbf{s}}$'' to ask ChatGPT to generate relational triplets even if it has low confidence in the triplet. We collect all the generated relational triplets from each sampled $\hat{\mathcal{S}}_i$ into a list $T_{\hat{\mathbf{s}}}$ and calculate the \textit{Demonstration Uncertainty} $U_{\hat{\mathbf{s}}}$ of each $T_i \in T_{\hat{\mathbf{s}}}$ as:
\begin{equation*}
    U_{\hat{\mathbf{s}}} = \left\{u_i\: \Big| \: 1-\frac{1}{N} \sum\nolimits_{j=1}^{N} \mathbbm{1}_{T_j=T_i}\right\},
\end{equation*}
%We quantify the variance for each extracted relation $T_i \in T_{\hat{\mathbf{s}}}$ by calculating its frequency and using a threshold $k$ to filter out low-frequent $T_i$. Formally, the final output relational triplet list $T_{\hat{\mathbf{s}}}$ is computed as:
where $N$ denotes the total number of elements in $T_{\hat{\mathbf{s}}}$ and $\mathbbm{1}$ is an indicator function counting each element's occurrence. Intuitively, if the uncertainty score $u_i$ is high, then it denotes the corresponding $T_i$ appears less frequent in $T_{\hat{\mathbf{s}}}$, and vice versa.

In this work, we adopt a threshold $k$ to filter high uncertain $T_i$, i.e., $T_{\hat{\mathbf{s}}} = \left\{T_i\: \Big| \: u_i \ge k\right\}$. Finally, based on the feedback from ChatGPT, we form the list of triplets $T_{\hat{\mathbf{s}}}$ to perform the evaluation.

\begin{table*}[t]
\resizebox{\textwidth}{!}{%
\begin{tabular}{@{}ll@{}}
\toprule
\textbf{Sentence}                                                                                & \begin{tabular}[c]{@{}l@{}}Although in Flanders, the Flemish Region assigned all of its powers to the Flemish Community, the \\ Walloon Region remains in principle distinct from and independent from the French Community, and vice-versa.\end{tabular}                                                                                                                                                                                                                                                             \\ \midrule
\textsc{GPT-3.5-Turbo}                                                                           & \begin{tabular}[c]{@{}l@{}}1. (the Flemish Region, remains, in principle distinct from and independent from the French Community)\\ 2. (Walloon Region, remains, in principle distinct from and independent from the French Community)\\ 3. (the French Community, remains, in principle distinct from and independent from the Walloon Region)\end{tabular}                                                                                                                                                          \\\midrule
\begin{tabular}[c]{@{}l@{}}\textsc{GPT-3.5-Turbo}\\ + \textsc{Selected Demo}\end{tabular}                 & \begin{tabular}[c]{@{}l@{}}1. (the Flemish Region, assigned, all of its powers to the Flemish Community)\\ 2. (the Walloon Region, remains, distinct from and independent from the French Community)\\ 3. (the Walloon Region, remains, in principle distinct from and independent from the French Community)\\ 4. (the French Community, is, distinct from and independent from the Walloon Region)\\ 5. (the French Community, is, in principle distinct from and independent from the Walloon Region)\end{tabular} \\\midrule
\begin{tabular}[c]{@{}l@{}}\textsc{GPT-3.5-Turbo}\\ + \textsc{Selected Demo}\\ + \textsc{Uncertainty}\end{tabular} & \begin{tabular}[c]{@{}l@{}}1. (the Flemish Region, assigned, all of its powers to the Flemish Community)\\ 2. (the Walloon Region, remains in principle distinct from, the French Community)\\ 3. (the Walloon Region, remains independent from, the French Community)\\ 4. (the French Community, is, distinct from and independent from the Walloon Region)\end{tabular}                                                                                                                                            \\\midrule\midrule
\textbf{Golden Standard}                                                                         & \begin{tabular}[c]{@{}l@{}}1. (the Flemish Region, assigned, all of its powers)\\ 2. (the Walloon Region, remains in principle distinct from, the French Community)\\ 3. (the Walloon Region, remains independent from, the French Community)\end{tabular} \\ \bottomrule
\end{tabular}%
}
\vspace{-3mm}
\caption{The extracted relational triplets with different demonstration strategies against the Golden Standard.}
\vspace{-5mm}
\label{tab: case}
\end{table*}

\section{Experiment}
The experiment is conducted with two family of LLMs, i.e., \textsc{LLaMA-2 13B \& 70B}\footnote{huggingface.co/meta-llama} and \textsc{GPT-3.5-turbo}\footnote{platform.openai.com/docs/models/gpt-3-5}. The framework is evaluated on three OIE benchmark datasets: (1) CaRB \cite{bhardwaj-etal-2019-carb}; (2) OIE2016 \cite{stanovsky2016creating}; and (3) ReOIE \cite{zhan2020span}. OIE systems are typically evaluated by comparing the extractions with the gold set in each dataset, and commonly used measures are Precision, Recall, and F1 scores. The F1 is the maximum value among the precision-recall pairs. We follow the matching function proposed for each dataset, i.e., lexical match for OIE2016 and ReOIE, and tuple match for CaRB.

\noindent\textbf{Baselines.} We adopt a list of recent OIE methods for comparison. (1) OpenIE4 \cite{mausam2016open} is a rule-based relational extraction method. (2) SpanOIE \cite{zhan2020span} is a supervised and span-based method that directly predicts whether a token span is an argument or a predicate. (3) IMOJIE \cite{kolluru2020imojie} is a generative OIE method that uses BERT as encoder and LSTM as decoder. Other than supervised methods, we incorporate different versions of our method with using different LLMs as the backbone model. Specifically, (4) \textsc{GPT-3.5-turbo} represents the vanilla version of the ChatGPT without any in-context learning and uncertainty quantification. (5) \textsc{LLaMA-2 w/ Fixed Demo} and \textsc{GPT-3.5-turbo w/ Fixed Demo} denote a fixed three sentences with annotated relational triplets are provided as a demonstration. (6) \textsc{LLaMA-2 w/ Selected Demo} and \textsc{GPT-3.5-turbo w/ Selected Demo} denotes the framework with customized in-context learning examples. (7) \textsc{LLaMA-2 w/ Selected Demo \& Uncertainty} and \textsc{GPT-3.5-turbo w/ Selected Demo \& Uncertainty} denotes the final framework with considering all components. Note that we adopt two versions of \textsc{LLaMA-2} model with different parameter size: \textsc{LLaMA-2-13B} and \textsc{LLaMA-2-70B}.

\subsection{Discussion}
The experiment results are depicted in Table \ref{tab:my-table}, and we draw a few conclusions from the result. First, the gap between the zero-shot LLM (\textsc{GPT-3.5-Turbo}) and the best methods across three datasets is around $30\%$ in F1 score across three datasets, which is reasonable since all SOTA methods are trained on corresponding datasets. Second, although few-shot LLM approachs are still not comparable to supervised fine-tuned methods in each dataset, the performance gap between the best LLM approach (i.e., \textsc{GPT-3.5-Turbo w/ Selected Demo \& Uncertainty}) and the supervised methods are very little (on average $6\%$ across three datasets). Third, with the growth of the parameter size, the few-shot OIE accuracy of LLMs also increases (\textsc{LLaMA-2-13B}$<$ \textsc{LLaMA-2-70B} $\approx$ \textsc{GPT-3.5-Turbo}). Even though we don't know the exact parameter size of \textsc{GPT-3.5-Turbo}, the performance of the state-of-the-art open source LLM \textsc{LLaMA-2-70B} is nearly the same as \textsc{GPT-3.5-Turbo} is the OIE task. We further investigate the effectiveness of different component in the following ablation study section.

\subsection{Ablation Study}
To mitigate the observed performance gap between the zero-shot LLMs and supervised methods, we propose several techniques to enhance the model's understanding towards the task and increase the trustworthiness of the prediction along the way. Taken the \textsc{GPT-3.5-Turbo} as an example, by adding a few demonstration examples, \textsc{GPT-3.5-Turbo w/ Fixed Demo} surpasses the zero-shot ChatGPT around $15\%$ in F1 score across all datasets. Furthermore, by customizing the demonstration examples and involving the uncertainty quantification module, the proposed framework can already achieve competitive and sometimes better results with the state-of-the-arts. The same trends are also observed in both \textsc{LLaMA-2-13B} and \textsc{LLaMA-2-70B} models.

\subsection{Case Study}
We further present a case study to showcase the effectiveness of each module, where we randomly pick a test instance in OIE2016 dataset and let each ablated model generate corresponding response. \textsc{GPT-3.5-Turbo} produced correct number of relations, but it incorrectly understood that the French Community was distinct and independent from the Walloon Region, which is not explicitly stated in the sentence. With adding selected demonstration examples, \textsc{GPT-3.5-Turbo+Selected Demo} provided more accurate interpretations but still over-interpreted the independence of the French Community from the Walloon Region. Finally, with considering the \textsc{GPT-3.5-Turbo+Selected Demo+Uncertainty} provided the most accurate interpretation compared to the golden standard.

\section{Conclusion}
We investigate the capability of LLMs being a zero/few-shot OIE system. We incorporate various in-context learning strategies to increase LLM's understanding towards the task and allow it generating structured output following the instruction. To further enhance the confidence of the generated output, we design a uncertainty quantification module to filter low-confident predictions. The proposed framework can achieve a competitive results with methods that are extensively trained on each dataset. This work can potentially serve as a starting point to freely mine entities and relations for constructing domain specific knowledge bases \cite{cui2023survey}.

\bibliographystyle{IEEEbib}
\bibliography{strings}

\begin{thebibliography}{10}

\bibitem{zhou2022survey}
Shaowen Zhou, Bowen Yu, Aixin Sun, Cheng Long, Jingyang Li, and Jian Sun,
\newblock ``A survey on neural open information extraction: Current status and
  future directions,''
\newblock {\em arXiv preprint arXiv:2205.11725}, 2022.

\bibitem{wang2018information}
Chengbin Wang, Xiaogang Ma, Jianguo Chen, and Jingwen Chen,
\newblock ``Information extraction and knowledge graph construction from
  geoscience literature,''
\newblock {\em Computers \& geosciences}, vol. 112, pp. 112--120, 2018.

\bibitem{ling2023knowledge}
Chen Ling, Xuchao Zhang, Xujiang Zhao, Yifeng Wu, Yanchi Liu, Wei Cheng,
  Haifeng Chen, and Liang Zhao,
\newblock ``Knowledge-enhanced prompt for open-domain commonsense reasoning,''
\newblock {\em AAAI Workshop on Uncertainty Reasoning and Quantification in
  Decision Making}, 2023.

\bibitem{chowdhury2023knowledge}
Tanmoy Chowdhury, Chen Ling, Xuchao Zhang, Xujiang Zhao, Guangji Bai, Jian Pei,
  Haifeng Chen, and Liang Zhao,
\newblock ``Knowledge-enhanced neural machine reasoning: A review,''
\newblock {\em arXiv preprint arXiv:2302.02093}, 2023.

\bibitem{niklaus2018survey}
Christina Niklaus, Matthias Cetto, Andr{\'e} Freitas, and Siegfried Handschuh,
\newblock ``A survey on open information extraction,''
\newblock {\em arXiv preprint arXiv:1806.05599}, 2018.

\bibitem{ouyang2022training}
Long Ouyang, Jeffrey Wu, Xu~Jiang, Diogo Almeida, Carroll Wainwright, Pamela
  Mishkin, Chong Zhang, Sandhini Agarwal, Katarina Slama, Alex Ray, et~al.,
\newblock ``Training language models to follow instructions with human
  feedback,''
\newblock {\em Advances in Neural Information Processing Systems}, vol. 35, pp.
  27730--27744, 2022.

\bibitem{ling2023beyond}
Chen Ling, Xujiang Zhao, Jiaying Lu, Chengyuan Deng, Can Zheng, Junxiang Wang,
  Tanmoy Chowdhury, Yun Li, Hejie Cui, Tianjiao Zhao, et~al.,
\newblock ``Domain specialization as the key to make large language models
  disruptive: A comprehensive survey,''
\newblock {\em arXiv preprint arXiv:2305.18703}, 2023.

\bibitem{yu2021maximal}
Bowen Yu, Yucheng Wang, Tingwen Liu, Hongsong Zhu, Limin Sun, and Bin Wang,
\newblock ``Maximal clique based non-autoregressive open information
  extraction,''
\newblock in {\em Proceedings of the 2021 Conference on Empirical Methods in
  Natural Language Processing}, 2021, pp. 9696--9706.

\bibitem{zhan2020span}
Junlang Zhan and Hai Zhao,
\newblock ``Span model for open information extraction on accurate corpus,''
\newblock in {\em Proceedings of the AAAI Conference on Artificial
  Intelligence}, 2020, vol.~34, pp. 9523--9530.

\bibitem{kolluru2020imojie}
Keshav Kolluru, Samarth Aggarwal, Vipul Rathore, Soumen Chakrabarti, et~al.,
\newblock ``Imojie: Iterative memory-based joint open information extraction,''
\newblock {\em arXiv preprint arXiv:2005.08178}, 2020.

\bibitem{lu2023pivoine}
Keming Lu, Xiaoman Pan, Kaiqiang Song, Hongming Zhang, Dong Yu, and Jianshu
  Chen,
\newblock ``Pivoine: Instruction tuning for open-world information
  extraction,''
\newblock {\em arXiv preprint arXiv:2305.14898}, 2023.

\bibitem{han2023information}
Ridong Han, Tao Peng, Chaohao Yang, Benyou Wang, Lu~Liu, and Xiang Wan,
\newblock ``Is information extraction solved by chatgpt? an analysis of
  performance, evaluation criteria, robustness and errors,''
\newblock {\em arXiv preprint arXiv:2305.14450}, 2023.

\bibitem{wei2023zero}
Xiang Wei, Xingyu Cui, Ning Cheng, Xiaobin Wang, Xin Zhang, Shen Huang, Pengjun
  Xie, Jinan Xu, Yufeng Chen, Meishan Zhang, et~al.,
\newblock ``Zero-shot information extraction via chatting with chatgpt,''
\newblock {\em arXiv preprint arXiv:2302.10205}, 2023.

\bibitem{su2022one}
Hongjin Su, Jungo Kasai, Yizhong Wang, Yushi Hu, Mari Ostendorf, Wen-tau Yih,
  Noah~A Smith, Luke Zettlemoyer, Tao Yu, et~al.,
\newblock ``One embedder, any task: Instruction-finetuned text embeddings,''
\newblock {\em arXiv preprint arXiv:2212.09741}, 2022.

\bibitem{bhardwaj-etal-2019-carb}
Sangnie Bhardwaj, Samarth Aggarwal, and Mausam Mausam,
\newblock ``{C}a{RB}: A crowdsourced benchmark for open {IE},''
\newblock in {\em Proceedings of the 2019 Conference on Empirical Methods in
  Natural Language Processing}, 2019.

\bibitem{stanovsky2016creating}
Gabriel Stanovsky and Ido Dagan,
\newblock ``Creating a large benchmark for open information extraction,''
\newblock in {\em Proceedings of the 2016 Conference on Empirical Methods in
  Natural Language Processing}, 2016, pp. 2300--2305.

\bibitem{mausam2016open}
Mausam Mausam,
\newblock ``Open information extraction systems and downstream applications,''
\newblock in {\em Proceedings of the twenty-fifth international joint
  conference on artificial intelligence}, 2016, pp. 4074--4077.

\bibitem{cui2023survey}
Hejie Cui, Jiaying Lu, Shiyu Wang, Ran Xu, Wenjing Ma, Shaojun Yu, Yue Yu, Xuan
  Kan, Chen Ling, Joyce Ho, et~al.,
\newblock ``A survey on knowledge graphs for healthcare: Resources,
  applications, and promises,''
\newblock {\em arXiv preprint arXiv:2306.04802}, 2023.

\end{thebibliography}

\end{document}